\documentclass[12pt, english]{article}

\usepackage{hyperref}
\hypersetup{pagebackref=true,breaklinks=true,letterpaper=true,colorlinks,bookmarks=false}
\usepackage{proposal}
\usepackage{geometry}
\usepackage{times}
\usepackage[T1]{fontenc}
\usepackage{epsfig}
\usepackage{graphicx}
\usepackage{amsmath}
\usepackage{amssymb}
\usepackage{array}
\usepackage[normalem]{ulem}
\usepackage{todo} 
\usepackage{xcolor}
\usepackage{subcaption}
\usepackage[printonlyused]{acronym}
\usepackage{algorithm}
\usepackage{algpseudocode} 
\usepackage{multirow}
\usepackage{booktabs}
\usepackage{mathtools}
\usepackage{enumitem}

\acrodef{KD}{Knowledge Distillation}
\acrodef{DL}{Deep Learning}

\newlength\savewidth

\graphicspath{{figures/}}
\date{}

\title{Data Efficient Stagewise Knowledge Distillation\\
}

\author{
	Akshay Kulkarni\\
	VNIT\\
	\small{\texttt{akshayk.vnit@gmail.com}}
	\and
	Navid Panchi\\
	VNIT\\
	\small{\texttt{navidpanchi@gmail.com}}
	\and
	Sharath Chandra Raparthy\\
	Mila\\
	\small{\texttt{raparths@mila.quebec}}
	\and
	Shital S. Chiddarwar\\
	VNIT\\
	\small{\texttt{shitalsc@mec.vnit.ac.in}}
}

\begin{document}
\maketitle

\begin{abstract}
Despite the success of \ac{DL}, the deployment of modern \ac{DL} models, which require high computational power, poses a significant problem for resource-constrained systems. This necessitates building compact networks that reduce computations while preserving performance. Traditional \ac{KD} methods that transfer knowledge from teacher to student \textbf{(a)} use a single-stage and \textbf{(b)} require the whole data set while distilling the knowledge to the student. In this work, we propose a new method called \textit{Stagewise Knowledge Distillation (SKD)} which builds on traditional KD methods by progressive stagewise training to leverage the knowledge gained from the teacher, resulting in data-efficient distillation process. We evaluate our method on classification and semantic segmentation tasks. We show, across the tested tasks, significant performance gains even with a fraction of the data used in distillation, without compromising on the metric. We also compare our method with existing KD techniques and show that SKD outperforms them. Moreover, our method can be viewed as a generalized model compression technique that complements other model compression methods such as quantization or pruning. The code is available at \url{https://github.com/IvLabs/stagewise-knowledge-distillation}.
\end{abstract}


\section{Introduction}
\label{sec:introduction}

Deep Learning has seen a huge success in the fields of Computer Vision\cite{NIPS2012_4824}, Robotics\cite{sunderhauf2018limits}, Natural Language Processing\cite{transformers}, etc. Over time, more complex models like ResNets\cite{resnet}, DenseNets\cite{densenet}, Inception Networks\cite{Szegedy_2015_CVPR}, etc., have been proposed at the cost of computation time and requirement of high-end GPUs. This does not matter when the application is run on the cloud or PCs with very high computing power with multiple GPUs. But, for smaller and more portable devices, it poses a computational bottleneck for real-time inference. Lack of sufficient computational power leads to long inference times, which poses a major limitation to real-time applications. For example, for speech-to-speech translation, very large Transformers\cite{transformers} are used, but these models cannot be deployed on embedded devices due to computational and memory constraints. Even if deployed, the inference time would render the application unusable. Since many of these applications are geared towards mobile and portable devices, this necessitates the use of more compact or computationally efficient networks. 

While a reduction in accuracy generally accompanies the use of smaller and compact networks due to low model capacity, several techniques have been developed which aim to maintain the accuracy while reducing the computational cost. These come under the research field of Model Compression in Deep Learning. Model compression techniques\cite{model_compression_review} can be broadly classified into the following  categories: 
\begin{enumerate}
    \item \textbf{Parameter Pruning and Sharing} approach reduces the redundancy in the parameters of the network which do not contribute to model performance
    \cite{hooker2019selective},  \cite{DBLP:journals/corr/SrinivasB15}, and  \cite{DBLP:journals/corr/HanPTD15}.
    \item \textbf{Low Rank Factorization techniques} aims to use tensor/matrix decomposition to determine useful parameters of the network \cite{DBLP:journals/corr/DentonZBLF14} and \cite{DBLP:journals/corr/JaderbergVZ14}.
    
    \item \textbf{Transferred/Compact Convolutional Filters} transforms over parameterized  filters to compact models resulting in computational efficiency \cite{DBLP:journals/corr/CohenW16}, \cite{DBLP:journals/corr/ShangSAL16}, \cite{DBLP:journals/corr/LiOW16}.
    \item \textbf{Knowledge Distillation} technique aims to train a compact model (student) using a larger pre-trained model (teacher) \cite{Hinton2015DistillingTK}, \cite{NIPS2014_5484} and \cite{structured_kd}.
    \item \textbf{Quantization and Binarization} aims to reduce the number of bits representing each weight of the network while preserving the network performance \cite{DBLP:journals/corr/GongLYB14}, \cite{DBLP:journals/corr/WuLWHC15}, and \cite{37631}.
\end{enumerate}
In this work, we focus on the knowledge distillation approach where we consider two models: \textbf{(a)} \textit{teacher} and \textbf{(b)} \textit{student}. Ideally, the teacher should be able to transfer all of its knowledge to the student, but this might not always be the case. Further, all of the teacher's knowledge won't necessarily be relevant to the student. In the best-case scenario, the student learns the most important details while leaving out the things which will not affect their performance in a given task. The teacher model is generally a larger model or an ensemble of models, whereas the student model is smaller and compact. Knowledge distillation techniques thus focus on training the student to mimic the output of the teacher model. 

Unlike previous KD methods \cite{Hinton2015DistillingTK}, \cite{fitnets}, we present a novel approach to train the student model using multiple feature maps from the teacher taken after particular layers. More specifically, the student model is trained in a \textit{stagewise} manner for each feature map, and the final classification layers are trained directly on the dataset without the teacher. We study the results of the proposed techniques for standard image classification and semantic segmentation tasks.  We show that this method enables the student to learn even on a small subset of the dataset. Note that, the aim of this work is not to improve the state-of-the-art on popular datasets, but to ensure that the discrepancy between the student and the teacher's accuracy is as low as possible.

\textit{\textbf{Contributions}}: We present a new approach, \textit{Stagewise Knowledge Distillation (SKD)}, for distilling the knowledge between teacher and student by using stagewise progressive training. Since the proposed method consists of training the student network one block at a time, the number of parameters updated at a given point in the distillation process is significantly smaller. As the student leverages the knowledge from the teacher while training, we explore the possibility of using a fraction of dataset for student training to achieve "data efficiency". To this end, we show that our method performs well over different image classification and semantic segmentation tasks. We also compare our method with existing KD techniques and show that SKD outperforms them.

\section{Related Work}
\label{sec:related_work}

\label{sec:lit_rev}
One of the first published works in model compression \cite{bucilua_modelcompression}, showed that knowledge from an ensemble could be transferred to a smaller model, which is trained to achieve performance similar to the ensemble. \cite{NIPS2014_5484} presented the first approach for knowledge distillation in deep neural networks. They achieve distillation by minimizing the MSE loss between the output of the teacher and student models. In this work, we use a similar approach except that we optimize the MSE Loss progressively stagewise. \cite{Hinton2015DistillingTK} extended the work of \cite{NIPS2014_5484} by using a combination of softened output from the teacher and the ground truth labels while training the student. In contrast, we directly make use of the ground truth labels and the intermediate feature maps making it more robust. Our method thus enables the student to learn the correct labels while resembling the feature maps of the teacher model. Our approach can be viewed as a generalization of \cite{fitnets}, in which the feature map from the middle layer of a large pre-trained teacher model is used to train the smaller student model along with the data. But in this work, we explore the possibility of using a fraction of the dataset, thereby reducing the distillation time.

\cite{gift_from_kd} proposed the use of feature maps to calculate a \textbf{F}low of \textbf{S}olution \textbf{P}rocedure (FSP) matrix and minimize the difference between the FSP matrices of the student and teacher models. However, we directly minimize the difference between the multiple feature maps of the student and the teacher, omitting the computationally expensive FSP matrix calculation. \cite{improved_kd_ta} introduced the idea of using an intermediate sized teaching assistant model between the models. Though this improves knowledge transfer, it introduces the additional memory and computational burden in training the student model. \cite{atkd} combined the idea of visual attention mapping with knowledge distillation techniques. They perform distillation on the attention maps instead of the feature maps themselves, with the motivation that attention maps are enough to convey the important information. However, the attention mapping functions are not learnable and thus handcrafted while also introducing additional computations.

Knowledge Distillation techniques have been successfully applied to semantic segmentation tasks. \cite{improving_fast_seg} uses the final outputs of the student and teacher networks to distill the knowledge using a consistency norm. They also use unlabelled data where the output of the teacher network is considered as ground truth for the student network. For dense prediction problems, \cite{structured_kd} proposed pairwise distillation that distills the pairwise similarities by building a static graph and holistic distillation using adversarial training. \cite{segkd_incremental} proposed the use of knowledge distillation techniques to incrementally learn new classes in semantic segmentation. However, the work focuses on learning new classes without reducing performance on existing classes,  which leans towards continual learning, instead of purely compressing the model. In this work, we present an approach that directly minimizes the distance between intermediate feature maps in a stagewise manner.

\begin{figure}[h!]
\includegraphics[width=15cm,height=15cm,keepaspectratio]{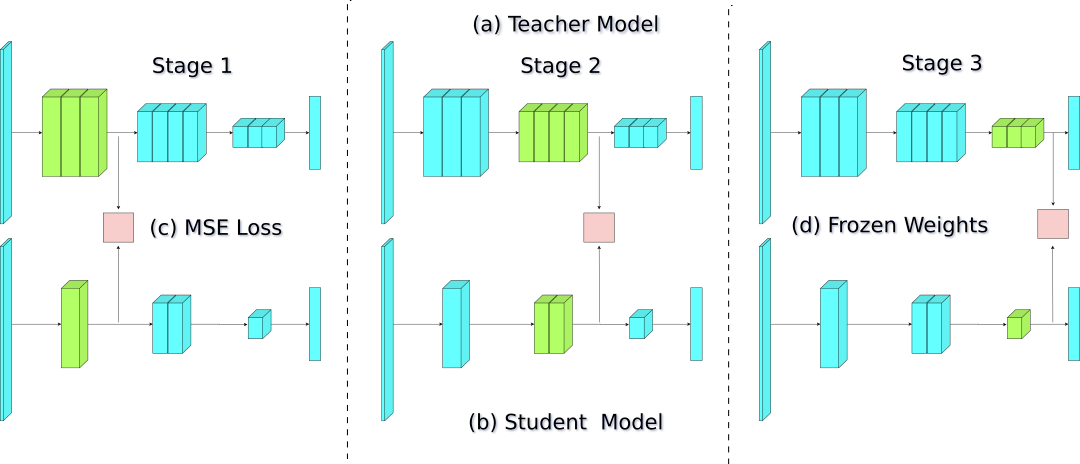}
\centering
\caption{Stagewise Knowledge Distillation approach aims to train a compact \textbf{student model} (b) to mimic the \textbf{teacher model} (a) by minimizing the \textbf{Mean Squared Error (MSE) Loss} at each stage(c) one at a time while \textbf{freezing the parameters}(d) of the other stages. Shown is the schematic which captures the higher level intuition of the method but not the specific architectural details. Blue represents frozen parameters while green represents trainable parameters.}
\label{fig:network_arch}
\end{figure}

\section{Methodology}
\label{sec:methodology}

We propose a simple yet powerful method for efficient knowledge distillation, \textit{Stagewise Knowledge Distillation (SKD)}. The approach is graphically shown in Figure \ref{fig:network_arch}. We consider a teacher model $f_\theta : \mathcal{X} \rightarrow \mathcal{Y}$ which is pretrained on a dataset $\mathcal{D}$ and a compact student model $f_\phi : \mathcal{X} \rightarrow \mathcal{Y}$ which is trained to mimic the teacher. With the pretrained teacher model in hand, we pass the input to both teacher and student model and extract the feature maps at the corresponding stages. We optimize the student model to reduce the MSE given by
\begin{equation}
\label{eqn:stagewise_mse}
    \mathcal{L}_2(y_{\theta}, y_{\phi}) = \frac{1}{\mathcal{M}}\sum_{j = 1}^\mathcal{M}||y_{\theta}(i, j) - y_{\phi}(i, j)||_2^2
\end{equation}
where $||.||_2$ represents $l_2$-norm, $y_\theta(i, j)$ is the intermediate output of the $i^{th}$ block corresponding to the $j^{th}$ input of the batch, of the teacher model. $y_\phi(i, j)$ is the intermediate output of the $i^{th}$ block corresponding to the $j^{th}$ input of the batch of the student model. $\mathcal{M}$ is the batch size. 

\begin{algorithm}[h!]
	\caption{Stagewise Knowledge Distillation}
	\begin{algorithmic}[1]
	\State \textbf{Required} Trained teacher model: $f_{\theta} : \mathcal{X} \rightarrow \mathcal{Y}$
		\State \textbf{Required:} Student Model: $f_{\phi} : \mathcal{X} \rightarrow \mathcal{Y}$
	    \State \textbf{Define:} MSE Loss $\mathcal{L}_2(y, \hat{y})$ using Eq \textbf{\ref{eqn:stagewise_mse}}
	    \State \textbf{Define:} Cross Entropy Loss $\mathcal{H}(y, \hat{y})$ using Eq \textbf{\ref{eqn:stagewise_ce}}
		\For {each stage \textit{s}}
		\State Freeze the parameters of other blocks of $f_\phi$ except for stage \textit{s}
    		\For {each batch \textit{x}}
        		\State Pass \textit{x} through  $f_\theta$, $f_\phi$ and get the feature maps $y_{\theta}^s$, $y_{\phi}^s$  for stage \textit{s} 
        		\State Calculate MSE Loss for stage \textit{s}: $\mathcal{L}^s$ =  $\mathcal{L}_2(y_{\theta}^s, y_{\phi}^s)$
        		\State Update the student model $f_{\phi}$
        		\EndFor
        	\EndFor 
        \State Train the classifier
    	\For {each batch \textit{x}}
    	    \State Pass \textit{x} through Student Model $y$ =  $f_{\phi}(\textit{x})$
    	    \State Calculate Cross Entropy Loss $\mathcal{L}$ = $\mathcal{H}(y, \hat{y})$
    	    \State Update the student model $f_{\phi}$
    		
    	    \EndFor
	\end{algorithmic} 
	\label{alg:skd}
\end{algorithm}

We perform this procedure sequentially in a block-by-block manner while freezing the parameters in the other blocks and hence optimizing only a subset of parameters at a time. In this way, ideally, the student is trained to mimic the teacher at each stage independently. In the last stage, the classifier part of the student is independently trained without a teacher to predict the classes using the standard Cross Entropy Loss given as follows:  
\begin{equation}
\label{eqn:stagewise_ce}
    \mathcal{H}(y, \hat{y}) = \frac{1}{\mathcal{M}}\sum_{j = 1}^\mathcal{M}\left\{-log\left(\frac{\exp({y_c(j, c))}}{\sum_{k = 1}^C \exp(y_c(j, k))}\right)\right\}
\end{equation}
where $y_c(j, k)$ is the output of the model corresponding to the $j^{th}$ input of the batch and the $k^{th}$ class, $\mathcal{C}$ is the number of classes, $c$ is the correct class (label) for a particular input and the remaining notations are same as in Eq \ref{eqn:stagewise_mse}. The overall training procedure is summarized in Algorithm \ref{alg:skd}. 

The stagewise training has various advantages, with a major one being the limited number of parameters that need to be optimized at a time. This leads to less strictness compared to training more number of parameters at once.  One more advantage of using the stagewise approach is the data efficiency. Since the teacher is trained on the whole dataset, student learning can be viewed in two ways:

\begin{enumerate}[label=(\alph*)]
    \item knowledge gained by mimicking the teacher at every stage.
    \item knowledge gained during the training of the classifier.
\end{enumerate}

In part \textbf{(a)}, the student relies entirely on the teacher for gaining the knowledge, while in part \textbf{(b)}, the student can learn independently of the teacher. Hence the student ideally leverages both of these as much as possible while still learning from less data and achieving the same accuracy. Also, from the training time perspective, it is useful to perform stagewise training using only a subset of the data to reduce training time. 

\section{Experimental Setup}
In this section we detail the experimental setup. We consider two tasks namely (a) Image Classification and (b) Semantic Segmentation. 

\subsection{Datasets}
For the image classification tasks, we use three  datasets: Imagenette\cite{imagenet_cvpr09}, Imagewoof\cite{imagenet_cvpr09} and CIFAR10\cite{cifar}. The first two datasets are subsets of the ImageNet\cite{imagenet} dataset which only vary in terms of difficulty. The former is an easier dataset for classification while the latter is relatively difficult. For the semantic segmentation task, we use CamVid\cite{camvid} dataset to test our approach. The classes used for training are diverse. Some of them include sky, building, pole, road, pavement, tree, sign-symbol, fence, car, pedestrian, and bicyclist. The dataset splits are given in Table \ref{tab:datasets}.

\begin{table}[h!]
\centering
\caption{Number of examples in the datasets}
\label{tab:datasets}
\begin{tabular}{cccc}
\toprule
 & Training & Validation & Testing \\ \midrule
Imagenette & 13000 & 500 & NA \\ \midrule
Imagewoof & 13000 & 500 & NA \\ \midrule
CIFAR10 & 50000 & 10000 & NA \\ \midrule
CamVid & 367 & 101 & 233 \\ \bottomrule
\end{tabular}
\end{table}

\subsection{Evaluation Metrics}
\subsubsection{Image Classification}
We use top-1 validation accuracy as an evaluation metric for classification tasks. It is the ratio of the number of correct predictions to the total number of predictions. Since all the datasets used are reasonably balanced, it is appropriate to report and compare using this metric.

\subsubsection{Semantic Segmentation} 
We report the mean Intersection over Union (IoU) over all classes on the validation set for the semantic segmentation task. IoU can be defined as the ratio of intersection of prediction and the ground truth to the union of the prediction and the ground truth segmentation.

\subsection{Baselines}
We compare our method against the following baselines and their variants:
\begin{itemize}
    \item \textbf{No Teacher:} The student models are trained without any teacher network, hence, there is no knowledge distillation. This baselines helps us in evaluation of improvement in different metrics for different Knowledge distillation methods. 
    \item \textbf{Simultaneous KD:} In Simultaneous KD, we jointly optimize the student to mimic the teacher and classify the labels. So,  MSE between each pair of feature maps is added together. Further, the cross-entropy loss between softmax output from the model and the label is added to the sum of the MSE losses. Thus, the loss can be represented as 
\begin{equation}
\label{eqn:simultaneous_loss}
    \begin{aligned}
        \mathcal{L}(y_\theta, y_\phi, y_c, c) = \frac{1}{\mathcal{MN}}\sum_{i = 1}^\mathcal{N}\sum_{j = 1}^\mathcal{M} ||y_\theta(i, j) - y_\phi(i, j)||_2^2 \\ + \frac{1}{\mathcal{M}}\sum_{j = 1}^\mathcal{M}\left\{-log\left(\frac{\exp(y_c(j, c))}{\sum_{k = 1}^C\exp(y_c(j, k))}\right)\right\} 
    \end{aligned}
\end{equation}
    where $N$ is the number of blocks and the remaining notations are the same as those used in Eq \ref{eqn:stagewise_mse} and \ref{eqn:stagewise_ce}.
    \item \textbf{Traditional KD:} As we can see that \textit{No Teacher} is a weak baseline. So, we also compare our method with the knowledge distillation method "FitNets" \cite{fitnets}. Hereafter, we refer to this baseline as "Traditional KD". In this, the student is first trained to mimic one intermediate feature map of the teacher. After this, the student is trained to perform the particular task (either classification or segmentation). 
    \item \textbf{FSP KD:} This method called "Flow of Solution Procedure" (FSP) \cite{gift_from_kd} attempts to teach the student network a flow of solving the problem instead of mimicking the teacher network's intermediate outputs. This is done by training the student to mimic the FSP matrices of the teacher. An FSP matrix is the inner product of feature maps output by 2 different layers of the network. In this, the student is first trained to mimic three FSP matrices (as per the original paper). Following this, the student is trained to perform the particular task (either classification or segmentation).
    \item \textbf{Attention Transfer KD (AT KD):} This technique, introduced in \cite{atkd}, uses visual attention mapping on intermediate feature maps and distills the knowledge by training the student to mimic the attention maps of the teacher. In these experiments, as per the original paper, the student is trained to mimic multiple attention maps of the teacher while also learning the main task (either classification or segmentation). 
\end{itemize}

\subsection{Image Classification Networks}
For the image classification experiments, several variants of deep residual networks\cite{resnet} are used. The models used are detailed in Table \ref{table:models}.
\subsubsection{Teacher Network} 
The teacher network is generally a standard model like ResNet18 or ResNet34. This makes it easier to obtain pre-trained weights for standard datasets. In this work, ResNet34 is used as the teacher model. 

\subsubsection{Student Network}
A student model is a smaller yet similar version of a certain standard teacher model. In this work, the student models are based on the ResNet34 model. They have fewer layers compared to the ResNet34 model (the first two layers having the 7x7 convolutional layer and max-pooling layer remain unchanged). However, the structure of the teacher network is preserved in the student networks, all of which are detailed in Table \ref{table:models}.

\begin{table}[h!]
\centering
\caption{Description of architectures used in this work}
\begin{tabular}{cccccc}
\toprule
 & \begin{tabular}[c]{@{}c@{}}ResNet34 \\ (standard)\end{tabular} & ResNet26 & ResNet20 & ResNet14 & ResNet10 \\ \toprule
conv1 & \multicolumn{5}{c}{7x7, 64, stride 2} \\ \midrule
\multirow{2}{*}{conv2\_x} & \multicolumn{5}{c}{3x3, maxpool, stride 2} \\ \cmidrule{2-6} 
 & $\begin{bsmallmatrix} 3\times3, 64 \\ 3\times3, 64 \end{bsmallmatrix}\times3$ & $\begin{bsmallmatrix} 3\times3, 64 \\ 3\times3, 64 \end{bsmallmatrix}\times3$ & $\begin{bsmallmatrix} 3\times3, 64 \\ 3\times3, 64 \end{bsmallmatrix}\times2$ & $\begin{bsmallmatrix} 3\times3, 64 \\ 3\times3, 64 \end{bsmallmatrix}\times1$ & $\begin{bsmallmatrix} 3\times3, 64 \\ 3\times3, 64 \end{bsmallmatrix}\times1$ \\ \midrule
conv3\_x & $\begin{bsmallmatrix} 3 \times 3, 128 \\ 3 \times 3, 128 \end{bsmallmatrix} \times 4$ & $\begin{bsmallmatrix} 3 \times 3, 128 \\ 3 \times 3, 128 \end{bsmallmatrix} \times 3$ & $\begin{bsmallmatrix} 3 \times 3, 128 \\ 3 \times 3, 128 \end{bsmallmatrix} \times 2$ & $\begin{bsmallmatrix} 3 \times 3, 128 \\ 3 \times 3, 128 \end{bsmallmatrix} \times 1$ & $\begin{bsmallmatrix} 3 \times 3, 128 \\ 3 \times 3, 128 \end{bsmallmatrix} \times 1$ \\ \midrule
conv4\_x & $\begin{bsmallmatrix} 3 \times 3, 256 \\ 3 \times 3, 256 \end{bsmallmatrix} \times 6$ & $\begin{bsmallmatrix} 3 \times 3, 256 \\ 3 \times 3, 256 \end{bsmallmatrix} \times 3$ & $\begin{bsmallmatrix} 3 \times 3, 256 \\ 3 \times 3, 256 \end{bsmallmatrix} \times 3$ & $\begin{bsmallmatrix} 3 \times 3, 256 \\ 3 \times 3, 256 \end{bsmallmatrix} \times 2$ & $\begin{bsmallmatrix} 3 \times 3, 256 \\ 3 \times 3, 256 \end{bsmallmatrix} \times 1$ \\ \midrule
conv5\_x & $\begin{bsmallmatrix} 3 \times 3, 512 \\ 3 \times 3, 512 \end{bsmallmatrix} \times 3$ & $\begin{bsmallmatrix} 3 \times 3, 512 \\ 3 \times 3, 512 \end{bsmallmatrix} \times 3$ & $\begin{bsmallmatrix} 3 \times 3, 512 \\ 3 \times 3, 512 \end{bsmallmatrix} \times 2$ & $\begin{bsmallmatrix} 3 \times 3, 512 \\ 3 \times 3, 512 \end{bsmallmatrix} \times 2$ & $\begin{bsmallmatrix} 3 \times 3, 512 \\ 3 \times 3, 512 \end{bsmallmatrix} \times 1$ \\ \midrule
FLOPs & 3.679G & 2.752G & 2.056G & 1.359G & 896.197M \\ \midrule
Parameters & 21.550M & 17.712M & 12.622M & 11.072M & 5.171M \\ \bottomrule
\end{tabular}
\label{table:models}
\end{table}

\subsection{Semantic Segmentation Networks}
For the semantic segmentation experiments, several variants of U-Nets \cite{unet} are used. The ResNet variants mentioned in Table \ref{table:models} are used as the encoder, and the decoder is symmetrically constructed as described in the original U-Net paper. Similar to the classification experiments, a ResNet34 based U-Net is used as the teacher model, whereas other smaller ResNets are used as encoders to construct the student U-Nets. 
\subsection{Implementation Details}
All experiments were performed using a computer having Intel Core i7-7700K CPU with a single Nvidia GTX1080Ti GPU. PyTorch\cite{pytorch} and fastai\cite{fastai} are used for implementation, training, and evaluation of all experiments. The image classification experiments use Adam optimizer\cite{adam} with a learning rate of 1e-4 for 100 epochs for each stage of training (Simultaneous KD, Attention Transfer KD and training without teacher are each done for 100 epochs). Each stage of Traditional KD and FSP KD is also run for 100 epochs with the same settings. The semantic segmentation experiments use Adam optimizer and One Cycle learning rate scheduler \cite{onecyclelr} with maximum learning rate 1e-2 for 50 epochs for each stage of training (Simultaneous KD and training without a teacher are also done for 50 epochs). The code is open-sourced at \url{https://github.com/IvLabs/stagewise-knowledge-distillation}.


\begin{table}[h!]
    \centering
    \captionsetup{justification=centering}
    \caption{Validation Accuracy for Imagenette \\ Validation Accuracy of ResNet34 Teacher is \textbf{99.2\%}}
    \label{table:imagenette}
    \begin{tabular}{lccccc}
    \toprule
    & \ ResNet10 \ & \ ResNet14 \ & \ ResNet18 \ & \ ResNet20 \ & \ ResNet26 \  \\ \midrule
    No Teacher & \ 91.8 \ & \ 91.2 \ & \ 91.4 \ & \ 91.6 \ & \ 90.6 \  \\
    Traditional KD & \ 91.2 \ & \ 90.0 \ & \ 90.2 \ & \ 91.2 \ & \ 90.0 \  \\
    Simultaneous KD & \ 92.2 \ & \ 93.2 \ & \ 92.4 \ & \ 92.4 \ & \ 91.8 \  \\
    FSP KD & \ 94.2 \ & \ 95.6 \ & \ 94.4 \ & \ 94.2 \ & \ 93.6 \  \\
    AT KD & \ 92.0 \ & \ 90.4 \ & \ 91.4 \ & \ 91.0 \ & \ 90.2 \  \\
    Stagewise KD & \ \textbf{97.4} \ & \ \textbf{ 98.8 }\ & \ \textbf{98.8} \ & \ \textbf{98.8} \ & \ \textbf{99.0 }\  \\ \midrule
    
    No Teacher - 10\% Data & \ 80.4 \ & \ 79.0 \ & \ 76.4 \ & \ 77.0 \ & \ 74.8 \  \\
    Traditional KD - 10\% Data & \ 77.4 \ & \ 78.0 \ & \ 77.0 \ & \ 77.0 \ & \ 74.2 \  \\
    Simultaneous KD - 10\% Data & \ 75.2 \ & \ 76.6 \ & \ 77.2 \ & \ 74.6 \ & \ 76.0 \  \\
    FSP KD - 10\% Data & \ 78.4	\ & \ 76.6 \ & \ 73.8 \ & \ 72.8 \ & \ 73.0 \  \\
    AT KD - 10\% Data & \ 79.8 \ & \ 78.4 \ & \ 77.2 \ & \ 77.0 \ & \ 75.4 \  \\
    Stagewise KD - 10\% Data & \ \textbf{86.6} \ & \ \textbf{87.8} \ & \ \textbf{82.6}\ & \ \textbf{83.6} \ & \ \textbf{85.4} \  \\ 
    \midrule
    
    No Teacher - 20\% Data & \ 84.8 \ & \ 83.0 \ & \ 81.8 \ & \ 81.6 \ & \ 81.6 \  \\
    Traditional KD - 20\% Data & \ 86.6 \ & \ 86.0 \ & \ 86.4 \ & \ 84.8 \ & \ 85.6 \  \\
    Simultaneous KD - 20\% Data & \ 82.6 \ & \ 82.6 \ & \ 80.6 \ & \ 82.0 \ & \ 81.4 \  \\
    FSP KD - 20\% Data & \ 85.0 \ & \ 82.4 \ & \ 82.4 \ & \ 80.6 \ & \ 80.6 \  \\
    AT KD - 20\% Data & \ 84.2 \ & \ 84.4 \ & \ 82.4 \ & \ 83.4 \ & \ 81.6 \  \\
    Stagewise KD - 20\% Data & \ \textbf{91.6} \ & \ \textbf{94.4} \ & \ \textbf{94.8}\ & \ \textbf{92.4} \ & \ \textbf{94.2} \  \\ 
    \midrule
    
    No Teacher - 30\% Data & \ 85.8 \ & \ 84.8 \ & \ 87.2 \ & \ 86.2 \ & \ 85.0 \  \\
    Traditional KD - 30\% Data & \ 90.0 \ & \ 89.2 \ & \ 89.4 \ & \ 90.6 \ & \ 90.6 \  \\
    Simultaneous KD - 30\% Data & \ 85.8 \ & \ 86.0 \ & \ 86.6 \ & \ 85.8 \ & \ 85.6 \  \\
    FSP KD - 30\% Data & \ 88.6	\ & \ 88.8 \ & \ 85.6 \ & \ 86.2 \ & \ 83.8 \  \\
    AT KD - 30\% Data & \ 85.6 \ & \ 85.4 \ & \ 83.8 \ & \ 84.4 \ & \ 85.0 \  \\
    Stagewise KD - 30\% Data & \ \textbf{96.0} \ & \ \textbf{97.0} \ & \ \textbf{97.4}\ & \ \textbf{97.0} \ & \ \textbf{97.4} \  \\ \midrule
    
    No Teacher - 40\% Data & \ 86.6 \ & \ 86.4 \ & \ 86.8 \ & \ 86.2 \ & \ 85.6 \  \\
    Traditional KD - 40\% Data & \ 91.6 \ & \ 90.6 \ & \ 92.2 \ & \ 92.0 \ & \ 92.0 \  \\
    Simultaneous KD - 40\% Data & \ 87.8 \ & \ 86.8 \ & \ 87.6 \ & \ 87.0 \ & \ 86.0 \  \\
    FSP KD - 40\% Data & \ 89.6	\ & \ 89.2 \ & \ 88.8 \ & \ 88.8 \ & \ 86.0 \  \\
    AT KD - 40\% Data & \ 88.2 \ & \ 88.2 \ & \ 88.0 \ & \ 86.2 \ & \ 85.8 \  \\
    Stagewise KD - 40\% Data & \ \textbf{96.4} \ & \ \textbf{97.4} \ & \ \textbf{97.6}\ & \ \textbf{97.6} \ & \ \textbf{97.8} \  \\ 
    \bottomrule
  \end{tabular}
\end{table}

\begin{table}[h!]
    \centering
    \captionsetup{justification=centering}
    \caption{Validation Accuracy for Imagewoof \\ Validation Accuracy of ResNet34 Teacher is \textbf{91.4}\%}
    \label{table:imagewoof}
    \begin{tabular}{lccccc}
    \toprule
    & \ ResNet10 \ & \ ResNet14 \ & \ ResNet18 \ & \ ResNet20 \ & \ ResNet26 \  \\ \midrule
    No Teacher & \ 80.2 \ & \ 78.6 \ & \ 79.2 \ & \ 79.8 \ & \ 80.2 \  \\
    Traditional KD & \ 77.2 \ & \ 78.4 \ & \ 79.4 \ & \ 78.8 \ & \ 77.8 \  \\
    Simultaneous KD & \ 79.8 \ & \ 79.6 \ & \ 81.0 \ & \ 81.4 \ & \ 84.2 \  \\
    FSP KD & \ 83.0 \ & \ 84.0 \ & \ 84.8 \ & \ 83.8 \ & \ 83.0 \  \\
    AT KD & \ 79.2 \ & \ 79.6 \ & \ 79.0 \ & \ 79.6 \ & \ 77.2 \  \\
    Stagewise KD & \ \textbf{90.6} \ & \ \textbf{92.8} \ & \ \textbf{92.4} \ & \ \textbf{92.0} \ & \ \textbf{93.4} \  \\ \midrule
    
    No Teacher - 10\% Data & \ 48.2 \ & \ 45.2 \ & \ 43.8 \ & \ 46.0 \ & \ 42.8 \  \\
    Traditional KD - 10\% Data & \ 51.0 \ & \ 50.8 \ & \ 49.8 \ & \ 51.0 \ & \ 46.2 \  \\
    Simultaneous KD - 10\% Data & \ 45.6 \ & \ 46.4 \ & \ 47.6 \ & \ 46.2 \ & \ 47.8 \  \\
    FSP KD - 10\% Data & \ 50.6	\ & \ 45.4 \ & \ 46.6 \ & \ 46.4 \ & \ 41.2 \  \\
    AT KD - 10\% Data & \ 48.8 \ & \ 45.4 \ & \ 47.4 \ & \ 45.8 \ & \ 45.0 \  \\
    Stagewise KD - 10\% Data & \ \textbf{70.4} \ & \ \textbf{68.6} \ & \ \textbf{64.4}\ & \ \textbf{63.8} \ & \ \textbf{63.8} \  \\ 
    \midrule
    
    No Teacher - 20\% Data & \ 64.0 \ & \ 60.4 \ & \ 60.0 \ & \ 60.6 \ & \ 55.0 \  \\
    Traditional KD - 20\% Data & \ 70.6 \ & \ 67.4 \ & \ 71.6 \ & \ 71.4 \ & \ 72.8 \  \\
    Simultaneous KD - 20\% Data & \ 60.2 \ & \ 58.0 \ & \ 60.8 \ & \ 63.2 \ & \ 63.2 \  \\
    FSP KD - 20\% Data & \ 63.6	\ & \ 58.2 \ & \ 54.2 \ & \ 55.6 \ & \ 50.0 \  \\
    AT KD - 20\% Data & \ 63.0 \ & \ 57.8 \ & \ 57.0 \ & \ 58.4 \ & \ 54.8 \  \\
    Stagewise KD - 20\% Data & \ \textbf{83.4} \ & \ \textbf{86.2} \ & \ \textbf{86.8}\ & \ \textbf{84.6} \ & \ \textbf{86.2} \  \\ 
    \midrule
    
    No Teacher - 30\% Data & \ 66.8 \ & \ 63.4 \ & \ 62.8 \ & \ 63.4 \ & \ 61.6 \  \\
    Traditional KD - 30\% Data & \ 77.0 \ & \ 75.0 \ & \ 77.4 \ & \ 77.4 \ & \ 79.6 \  \\
    Simultaneous KD - 30\% Data & \ 63.2 \ & \ 64.0 \ & \ 64.4 \ & \ 66.0 \ & \ 65.4 \  \\
    FSP KD - 30\% Data & \ 71.2	\ & \ 69.2 \ & \ 63.4 \ & \ 66.6 \ & \ 61.0 \  \\
    AT KD - 30\% Data & \ 65.0 \ & \ 63.8 \ & \ 62.8 \ & \ 63.8 \ & \ 61.6 \  \\
    Stagewise KD - 30\% Data & \ \textbf{87.0} \ & \ \textbf{90.0} \ & \ \textbf{89.6}\ & \ \textbf{89.4} \ & \ \textbf{91.2} \  \\ \midrule
    
    No Teacher - 40\% Data & \ 69.0 \ & \ 70.0 \ & \ 66.6 \ & \ 69.0 \ & \ 63.2 \  \\
    Traditional KD - 40\% Data & \ 79.6 \ & \ 81.2 \ & \ 78.8 \ & \ 80.6 \ & \ 80.8 \  \\
    Simultaneous KD - 40\% Data & \ 69.6 \ & \ 68.0 \ & \ 69.0 \ & \ 71.8 \ & \ 71.0 \  \\
    FSP KD - 40\% Data & \ 75.8	\ & \ 75.8 \ & \ 72.4 \ & \ 72.2 \ & \ 69.8 \  \\
    AT KD - 40\% Data & \ 70.4 \ & \ 67.0 \ & \ 69.2 \ & \ 66.8 \ & \ 64.6 \  \\
    Stagewise KD - 40\% Data & \ \textbf{89.4} \ & \ \textbf{90.6} \ & \ \textbf{92.8}\ & \ \textbf{92.8} \ & \ \textbf{92.4} \  \\ 
    \bottomrule
  \end{tabular}
\end{table}

\begin{table}[h!]
    \centering
    \captionsetup{justification=centering}
    \caption{Validation Accuracy for CIFAR10 \\ Validation Accuracy of ResNet34 Teacher is \textbf{87.51}\%}
    \label{table:cifar10}
    \begin{tabular}{lccccc}
    \toprule
    & \ ResNet10 \ & \ ResNet14 \ & \ ResNet18 \ & \ ResNet20 \ & \ ResNet26 \  \\ \midrule
    No Teacher & \ 77.88 \ & \ 77.50 \ & \ 77.35 \ & \ 78.08 \ & \ 78.30 \  \\
    Traditional KD & \ 76.22 \ & \ 77.08 \ & \ 77.38 \ & \ 77.22 \ & \ 77.59 \  \\
    Simultaneous KD & \ 77.32 \ & \ 75.98 \ & \ 76.47 \ & \ 76.79 \ & \ 76.94 \  \\
    FSP KD & \ 77.92 \ & \ 77.41 \ & \ 77.46 \ & \ 77.67 \ & \ 78.11 \  \\
    AT KD & \ 78.25	\ & \ 78.18 \ & \ 78.08 \ & \ 78.14 \ & \ 78.27 \  \\
    Stagewise KD & \ \textbf{84.75} \ & \ \textbf{84.97} \ & \ \textbf{85.99} \ & \ \textbf{86.46} \ & \ \textbf{86.62} \  \\ \midrule
    
    No Teacher - 10\% Data & \ 57.45 \ & \ 56.22 \ & \ 55.53 \ & \ 54.76 \ & \ 53.05 \  \\
    Traditional KD - 10\% Data & \ 69.05 \ & \ 68.10 \ & \ 68.44 \ & \ 68.08 \ & \ 67.42 \  \\
    Simultaneous KD - 10\% Data & \ 54.67 \ & \ 53.68 \ & \ 52.78 \ & \ 51.89 \ & \ 52.35 \  \\
    FSP KD - 10\% Data & \ 56.73 \ & \ 56.19 \ & \ 54.98 \ & \ 55.8 \ & \ 54.64 \  \\
    AT KD - 10\% Data & \ 56.15 \ & \ 55.64 \ & \ 54.49 \ & \ 55.34 \ & \ 54.36 \  \\
    Stagewise KD - 10\% Data & \ \textbf{77.74} \ & \ \textbf{77.15} \ & \ \textbf{77.77}\ & \ \textbf{78.05} \ & \ \textbf{78.17} \  \\ \midrule
    
    No Teacher - 20\% Data & \ 63.29 \ & \ 62.26 \ & \ 61.55 \ & \ 61.99 \ & \ 61.44 \  \\
    Traditional KD - 20\% Data & \ 74.94 \ & \ 75.11 \ & \ 75.73 \ & \ 75.63 \ & \ 76.02 \  \\
    Simultaneous KD - 20\% Data & \ 61.79 \ & \ 59.90 \ & \ 59.42 \ & \ 59.54 \ & \ 59.57 \  \\
    FSP KD - 20\% Data & \ 64.56 \ & \ 64.72 \ & \ 63.26 \ & \ 62.65 \ & \ 63.97 \  \\
    AT KD - 20\% Data & \ 63.91	\ & \ 63.01 \ & \ 62.25 \ & \ 62.78 \ & \ 62.16 \  \\
    Stagewise KD - 20\% Data & \ \textbf{81.53} \ & \ \textbf{81.32} \ & \ \textbf{82.12}\ & \ \textbf{82.40} \ & \ \textbf{82.43} \  \\ \midrule
    
    No Teacher - 30\% Data & \ 67.32 \ & \ 66.64 \ & \ 66.37 \ & \ 66.90 \ & \ 66.52 \  \\
    Traditional KD - 30\% Data & \ 77.89 \ & \ 77.54 \ & \ 78.45 \ & \ 78.26 \ & \ 78.66 \  \\
    Simultaneous KD - 30\% Data & \ 65.18 \ & \ 64.00 \ & \ 64.77 \ & \ 63.94 \ & \ 64.52 \  \\
    FSP KD - 30\% Data & \ 67.26 \ & \ 67.19 \ & \ 66.96 \ & \ 66.57 \ & \ 68.19 \  \\
    AT KD - 30\% Data & \ 67.65	\ & \ 66.79 \ & \ 66.12 \ & \ 66.14 \ & \ 67.24 \  \\
    Stagewise KD - 30\% Data & \ \textbf{82.48} \ & \ \textbf{82.74} \ & \ \textbf{83.48}\ & \ \textbf{83.49} \ & \ \textbf{84.42} \  \\ \midrule
    
    No Teacher - 40\% Data & \ 69.48 \ & \ 69.05 \ & \ 69.42 \ & \ 69.54 \ & \ 69.64 \  \\
    Traditional KD - 40\% Data & \ 79.81 \ & \ 80.04 \ & \ 80.58 \ & \ 80.75 \ & \ 80.94 \  \\
    Simultaneous KD - 40\% Data & \ 67.57 \ & \ 67.44 \ & \ 67.65 \ & \ 67.37 \ & \ 68.16 \  \\
    FSP KD - 40\% Data & \ 70.03	\ & \ 69.47 \ & \ 69.29 \ & \ 69.82 \ & \ 69.81 \  \\
    AT KD - 40\% Data & \ 70.17	\ & \ 69.81 \ & \ 69.76 \ & \ 69.9 \ & \ 69.82 \  \\
    Stagewise KD - 40\% Data & \ \textbf{82.95} \ & \ \textbf{83.58} \ & \ \textbf{84.16}\ & \ \textbf{84.82} \ & \ \textbf{85.19} \  \\ 
    \bottomrule
  \end{tabular}
\end{table}

\begin{table}[h!]
    \centering
    \captionsetup{justification=centering}
    \caption{Validation IoU for CamVid dataset \\ Validation IoU for ResNet34 Teacher is \textbf{0.607}}
    \label{table:validation_iou}
    \begin{tabular}{lccccc}
    \toprule
    & \ ResNet10 \ & \ ResNet14 \ & \ ResNet18 \ & \ ResNet20 \ & \ ResNet26 \  \\ \midrule
    No Teacher & \ 0.526 \ & \ 0.547 \ & \ 0.542 \ & \ 0.538 \ & \ 0.559 \  \\
    Traditional KD & \ 0.595 \ & \ 0.596 \ & \ 0.593 \ & \ 0.605 \ & \ \textbf{0.602} \  \\
    Simultaneous KD & \ 0.583 \ & \ 0.577 \ & \ 0.579 \ & \ 0.555 \ & \ 0.599 \  \\
    Stagewise KD & \ \textbf{0.599} \ & \ \textbf{0.613} \ & \ \textbf{0.608} \ & \ \textbf{0.608} \ & \ 0.600 \  \\ \midrule
    
    No Teacher - 10\% Data & \ 0.259 \ & \ 0.268 \ & \ 0.267 \ & \ 0.236 \ & \ 0.260 \  \\
    Traditional KD - 10\% Data & \ 0.429 \ & \ 0.369 \ & \ 0.313 \ & \ 0.320 \ & \ 0.300 \  \\
    Simultaneous KD - 10\% Data & \ 0.401 \ & \ 0.405 \ & \ 0.386 \ & \ 0.405 \ & \ 0.397 \  \\
    Stagewise KD - 10\% Data & \ \textbf{0.527} \ & \ \textbf{0.545} \ & \ \textbf{0.549} \ & \ \textbf{0.551} \ & \ \textbf{0.545} \  \\ \midrule
    
    No Teacher - 20\% Data & \ 0.313 \ & \ 0.319 \ & \ 0.323 \ & \ 0.332 \ & \ 0.340  \  \\
    Traditional KD - 20\% Data & \ 0.475 \ & \ 0.470 \ & \ 0.464 \ & \ 0.471 \ & \ 0.472 \  \\
    Simultaneous KD - 20\% Data & \ 0.437 \ & \ 0.437 \ & \ 0.438 \ & \ 0.430 \ & \ 0.441 \  \\
    Stagewise KD - 20\% Data & \ \textbf{0.563} \ & \ \textbf{0.582} \ & \ \textbf{0.581} \ & \ \textbf{0.577} \ & \ \textbf{0.580} \  \\ \midrule
    
    No Teacher - 30\% Data & \ 0.366 \ & \ 0.380 \ & \ 0.374 \ & \ 0.367 \ & \ 0.365 \  \\
    Traditional KD - 30\% Data & \ 0.519 \ & \ 0.564 \ & \ 0.549 \ & \ 0.479 \ & \ 0.490 \  \\
    Simultaneous KD - 30\% Data & \ 0.472 \ & \ 0.479 \ & \ 0.486 \ & \ 0.465 \ & \ 0.470 \  \\
    Stagewise KD - 30\% Data & \ \textbf{0.570} \ & \ \textbf{0.597} \ & \ \textbf{0.591} \ & \ \textbf{0.587} \ & \ \textbf{0.587} \  \\ \midrule
    
    No Teacher - 40\% Data & \ 0.400 \ & \ 0.416 \ & \ 0.419 \ & \ 0.417 \ & \ 0.435 \  \\
    Traditional KD - 40\% Data & \ 0.555 \ & \ 0.552 \ & \ 0.478 \ & \ 0.450 \ & \ 0.438 \  \\
    Simultaneous KD - 40\% Data & \ 0.526 \ & \ 0.512 \ & \ 0.510 \ & \ 0.512 \ & \ 0.514 \  \\
    Stagewise KD - 40\% Data & \ \textbf{0.581} \ & \ \textbf{0.601} \ & \ \textbf{0.594} \ & \ \textbf{0.594} \ & \ \textbf{0.589} \  \\
    \bottomrule
  \end{tabular}
\end{table}

\section{Discussions}
\label{sec:results}
We show the efficacy of our method by comparing it against the baselines. Here we explicitly focus on the student's performance in comparison with the teacher and do not focus on achieving the state-of-the-art metrics for both the tasks. 

\subsection{Image Classification}
For the classification task, we perform experiments across multiple datasets to ensure that the proposed method generalizes well. We also performed experiments with a fraction of the data to show the effectiveness on limited data. From all three tables \ref{table:imagenette}, \ref{table:imagewoof}, \ref{table:cifar10}, it is evident that SKD achieves the least discrepancy between teacher and student models across all datasets, all baselines and all percentages of dataset used for training. In certain experiments with the Imagewoof dataset, the SKD trained student models even surpass the teacher model. 

In the experiments with the complete dataset, the other baselines (Traditional KD, AT KD, FSP KD and Simultaneous KD) sometimes fail to even surpass the No Teacher baseline. This can be attributed to the fact that these methods attempt to achieve multiple optimization objectives at the same time. For example, AT KD attempts to train the student to mimic multiple attention maps of the teacher model while also learning the main classification task. Our approach overcomes this difficulty by using a stagewise training method which also reduces the number of parameters being trained at a time hence reducing the strictness and constraints in the optimization.

Across different ResNet architectures, we observe that the performance of SKD generally increases for the full dataset experiments. In the less data regime, we observe a downward trend in validation accuracy from ResNet10 to ResNet26. This means that as the student model capacity increases, it overfits the training dataset. Another trend that is observed in less data (30\% and 40\%) baseline experiments is that validation accuracy first increases and then decreases from ResNet10 to ResNet26. The smaller models has lesser capacity and thus the performance improves as model size increases. However, since it decreases after a certain model size, it can be concluded that having more parameters causes overfitting and degrades the performance.It can be considered that the medium sized models are somewhere between too few parameters and too many parameters. As mentioned earlier, SKD performance increases with the student model capacity.  This reiterates the importance of stagewise training as SKD is less susceptible to overfitting since a fewer number of parameters are being trained at each stage.

The motivation for performing experiments using a fraction of the dataset is twofold. First, SKD only optimizes one stage at a time. Due to the less number of parameters being trained in any particular stage, the risk of overfitting is reduced and a reasonable performance is expected with the less data experiments. Further, as the teacher is trained using full data, the student network can leverage the teacher's knowledge by optimizing the mean squared loss per stage and thus we can constrain the number of samples we are using for student training. Second, using less training data implies faster training which is important since SKD takes longer to train compared to the baselines. However, it should be noted that the use of KD techniques is to generate a compact model to be deployed on resource-constrained hardware but the actual distillation can be performed on more powerful machines. Thus, the disadvantage of longer runtime is not very significant.

\subsection{Semantic Segmentation}

To ensure that the proposed approach is task agnostic, we also test it on semantic segmentation. We compare our approach with the No Teacher, Simultaneous KD and Traditional KD (FitNets) baselines. We present the results in Table 6. It can be seen that there is a minimum of 5\% increase in IoU over training without teacher and 1\% increase in IoU over Traditional KD and Simultaneous KD methods. In the \textit{less data} regime, without a teacher, the model performs very poorly as expected. However, when knowledge distillation is done using SKD, there is a huge increase in IoU. Surprisingly, it can be seen that models trained using SKD with 10\% of the dataset outperforms models trained using 40\% of the dataset without a teacher.

Since the amount of data in the CamVid dataset is very less compared to the datasets used for classification experiments, it is easier for the models to overfit the training data. Thus, the discrepancy between SKD and baselines for these experiments is not as large as in the image classification experiments.

\section{Conclusion and Future Work}
\label{sec:conclusion}
We propose a novel way to distill the knowledge from teacher to student model. The success of the proposed method can be attributed to the optimization of less number of parameters at a given stage which subdues the optimization difficulty of the traditional KD methods. We also explore the possibility of distillation in the \textit{less data} regime. We compare our method with other KD techniques and show that SKD outperforms them. The proposed method can be extremely helpful while performing the distillation on bigger datasets like ImageNet or Cityscapes \cite{cityscapes}.

Further, our method is flexible and can be used with other model compression techniques. We also show that our approach is task agnostic and can be applied across a wide range of tasks like segmentation and classification. While our method surpasses other KD techniques, it is possible to incorporate those methods in our approach to further improve performance. The scope of the proposed technique is boundless and it can be viewed as a generalized compression technique. We wish to further explore the applicability of the proposed method to other Computer Vision tasks such as object detection and pose estimation.

\section*{Acknowledgements}
The authors would like to thank all current and previous members of \href{http://www.ivlabs.in/}{IvLabs - the AI \& Robotics Club of VNIT} for their constant support and motivation. Special thanks to \href{https://prasadvagdargi.github.io/}{Prasad Vagdargi} and \href{https://scholar.google.com/citations?user=GdXOFKoAAAAJ&hl=en}{Berivan Isik} for proofreading the paper.


\end{document}